\documentclass{article}

\usepackage{arxiv}

\usepackage[utf8]{inputenc}
\usepackage[T1]{fontenc}

\usepackage{hyperref}
\usepackage{url}
\usepackage{booktabs}
\usepackage{amsfonts}
\usepackage{amsmath}
\usepackage{amssymb}
\usepackage{nicefrac}
\usepackage{microtype}
\usepackage{graphicx}
\usepackage{algorithm}
\usepackage{algpseudocode}

\graphicspath{ {./images/} }

\title{
From Language to Action in Arabic: Reliable Structured Tool Calling via Data-Centric Fine-Tuning}

\author{
    \begin{minipage}[t]{\textwidth}
        \centering
        \normalfont
        Omer Nacar\textsuperscript{1},
        Deema Alquffari\textsuperscript{1},
        Saleh Alsharideh\textsuperscript{1},
        Adeem AlOtaibi\textsuperscript{3},
        Abdulaziz Alabdulkarim\textsuperscript{2},
        Leen Alhazmi\textsuperscript{1},
        Nada Alomar\textsuperscript{1},
        Wareef Alzubaidi\textsuperscript{1},
        Nada Alsultan\textsuperscript{1},
        Ahmed Alrabghi\textsuperscript{1},
        Demah Alhoshan\textsuperscript{1},
        Rana Alsayyari\textsuperscript{5},
        Hamed Alruwaili\textsuperscript{1},
        Albaraa Jaafar\textsuperscript{4},
        Khaled Alusmani\textsuperscript{1},
        Abdulaziz Alsohimy\textsuperscript{1},
        Munirah Alsubaie\textsuperscript{3},
        Shahd Aldukhayil\textsuperscript{1},
        Arwa Alali\textsuperscript{1},
        Yazeed BinShihah\textsuperscript{1},
        Razan Alsulaymi\textsuperscript{1},
        Nourah Alhumaid\textsuperscript{1},
        Razan Abdulsalam\textsuperscript{1},
        Reem Alamoudi\textsuperscript{6},
        Mohammed Alkhalifa\textsuperscript{1} \\[1em]
        {
            \textsuperscript{1}Tuwaiq Academy, Riyadh, Saudi Arabia \\
            \textsuperscript{2}Tahakom, \textsuperscript{3}Wakeb Company, 
            \textsuperscript{4}Qatmeer Co.,
            \textsuperscript{5}NCGR,
            \textsuperscript{6}Vision Bank,
        } \\[0.5em]
        \texttt{\normalsize  o.najar@tuwaiq.edu.sa}
    \end{minipage}
}

\begin{document}
\maketitle

\begin{abstract}
Function-calling language models are essential for agentic AI systems that translate natural language into executable structured actions, yet existing models exhibit severe structural instability when applied to Arabic. We present \textbf{AISA-AR-FunctionCall}, a production-oriented Arabic function-calling framework built on a 270M-parameter FunctionGemma backbone and trained through systematic dataset auditing, schema repair, tool-aware prompt restructuring, and full-parameter supervised fine-tuning. On a held-out test set, fine-tuning reduces parse failures from 87\% to below 1\%, improves function name accuracy by more than eightfold, and substantially enhances argument alignment across dialects and domains. Error analysis reveals a transition from structural collapse to semantic misalignment, suggesting that serialization stability and decision-level reasoning are separable challenges. We further explore a reasoning-augmented LoRA variant that introduces explicit intermediate reasoning prior to tool invocation. All datasets and models are publicly released under the AISA framework.\footnote{\url{https://huggingface.co/collections/AISA-Framework/aisa-arabic-functioncall-datasets-and-models}}
\end{abstract}

\section{Introduction}

Large language models (LLMs) are increasingly deployed not merely as text generators, but as decision-making components in \emph{agentic} systems that translate natural language intent into executable actions. This capability—commonly referred to as \textbf{function calling} or \textbf{tool use}—sits at the boundary between language understanding and software execution. Instead of responding with free-form text, the model emits a structured representation of an API call, which an external runtime validates and executes before returning results to the model for final user-facing synthesis. Such patterns underpin modern assistants, enterprise workflow agents, and local-first automation systems \cite{yao2022react,karpas2022mrkl,schick2023toolformer}.

Despite rapid methodological progress, tool calling introduces new reliability and safety challenges. Failures often stem from malformed arguments, incorrect tool selection, schema violations, or brittle orchestration logic across multiple system layers. Importantly, these errors are rarely attributable to the base model alone; rather, they emerge from interactions between prompting formats, schema design, runtime validation, and evaluation blind spots \cite{patil2025bfcl,guo2024stabletoolbench}. As evaluation frameworks have matured—from ToolQA and ToolLLM to StableToolBench and BFCL—they reveal that structured execution remains substantially more difficult than text generation alone \cite{zhuang2023toolqa,qin2023toolllm,guo2024stabletoolbench,patil2025bfcl}.

Recent open releases aim to address format reliability directly. FunctionGemma, built on Gemma 3 270M, introduces a dedicated control-token interface for tool declaration, invocation, and response handling, alongside structured delimiters to reduce ambiguity between natural language and executable artifacts \cite{google2025functiongemma_overview,google2025functiongemma_formatting,hf_functiongemma_modelcard}. The Gemma family emphasizes lightweight, deployable architectures capable of specialization for on-device and privacy-preserving use cases \cite{gemma_team2024gemma2,gemma_team2025gemma3}. However, FunctionGemma is explicitly designed as a \emph{base} for domain- or language-specific fine-tuning, rather than as a production-ready multilingual agent out of the box.

A critical gap remains in \textbf{multilingual and Arabic tool-calling performance}. While multilingual benchmarks such as MASSIVE-Agents reformulate datasets across 52 languages and demonstrate standardized evaluation pipelines, they reveal substantial cross-lingual disparities in function-call correctness \cite{kulkarni2025massiveagents}. Performance outside English drops markedly even for strong multilingual models, reinforcing that structured execution does not transfer reliably across languages. In parallel, Arabic NLP research has produced strong localized language models—including AraBERT, ARBERT/MARBERT, Jais, and AceGPT \cite{antoun2020arabert,abdulmageed2021marbert,sengupta2023jais,huang2024acegpt}—yet tool-calling datasets and agentic evaluation resources in Arabic remain underdeveloped relative to English ecosystems.

This paper addresses that gap by presenting an Arabic-first function-calling dataset and a fully fine-tuned execution model, developed through a community-driven effort to localize and specialize FunctionGemma for Arabic structured action generation. We introduce (i) a large-scale Arabic dataset pairing natural-language requests with structured tool schemas and executable tool-call annotations, (ii) reasoning supervision in a \emph{reason-before-call} format inspired by chain-of-thought training \cite{wei2022cot,yao2022react}, and (iii) an evaluation protocol combining structure-level correctness metrics with Arabic-specific robustness tests for ambiguity, slot-filling, and refusal behavior.

Beyond dataset and model contributions, we frame the work as a \textbf{systems-level instantiation} grounded in \emph{AISA (Agentic AI Systems Architecture)} \cite{nacar2026aisa}. AISA separates concerns across foundational models, tool interfaces, orchestration infrastructure, evaluation layers, deployment controls, and governance mechanisms. Rather than treating function calling as a prompt-engineering artifact, we implement explicit cross-layer contracts: schema validation and safe dispatch at the tool layer, structured parsing and retry logic at the orchestration layer, versioned datasets and release gating at the deployment layer, and policy enforcement aligned with AI risk management guidance \cite{nist2023airmf,iso42001_2023,iso23894_2023}. This architecture-first perspective enables reproducible evaluation, auditability, and production-readiness for Arabic agentic systems.

In summary, our work positions Arabic tool calling at the intersection of three threads: (1) structured execution research in LLMs, (2) Arabic NLP localization and cultural alignment, and (3) governance-aware agent system engineering. By grounding Arabic function calling in both empirical fine-tuning and explicit architectural design, we aim to move from isolated model adaptation toward reliable, deployable Arabic agentic systems.

\section{Related Work}

The integration of external tools into language model reasoning has evolved from prompt-based experimentation to structured, format-controlled execution. Early paradigms such as ReAct interleaved reasoning traces with actions, demonstrating that explicit reasoning combined with external tool interaction improves task completion and interpretability \cite{yao2022react}. Similarly, MRKL systems proposed modular architectures in which language models route queries to symbolic or external components, emphasizing separation between reasoning and execution \cite{karpas2022mrkl}. Toolformer further showed that language models can self-supervise tool invocation behavior by learning when and how to call APIs during generation \cite{schick2023toolformer}.

As tool ecosystems expanded, large-scale datasets and benchmarks emerged to evaluate tool-use capabilities. ToolLLM introduced ToolBench, scaling training and evaluation to thousands of real-world APIs \cite{qin2023toolllm}. ToolQA focused specifically on question answering tasks requiring external tool usage rather than memorized knowledge \cite{zhuang2023toolqa}. StableToolBench emphasized stability and reproducibility through API simulation and caching mechanisms, highlighting evaluation brittleness in real API-dependent setups \cite{guo2024stabletoolbench}. UltraTool further expanded benchmarking to complex, real-world multi-step tool utilization scenarios \cite{huang2024ultratool}. The Berkeley Function Calling Leaderboard (BFCL) formalized structured evaluation of tool-calling correctness via abstract syntax tree (AST) comparisons and extended evaluation toward agentic behaviors \cite{patil2025bfcl}. However, most of these resources remain English-centric, leaving multilingual and morphologically rich languages underrepresented.

Reliable tool invocation depends on structured, parseable outputs. In practice, failures often arise from invalid JSON formatting, missing arguments, or schema mismatches. To mitigate such issues, recent systems adopt explicit structured-output enforcement. Closed APIs provide schema-constrained JSON generation and function-calling interfaces to ensure machine-readable outputs \cite{openai_function_calling_2025,openai_structured_outputs_2025,anthropic_tool_use_docs_2025}.

FunctionGemma represents an open-model effort toward format-controlled execution \cite{google2025functiongemma_overview,google2025functiongemma_formatting}. Built on Gemma 3 270M \cite{gemma_team2025gemma3}, it introduces six dedicated control tokens for tool lifecycle management (declaration, call, response) and a specialized delimiter token to disambiguate structured string fields. Importantly, the model card explicitly states that it is designed for specialization through fine-tuning and supports single-turn and parallel tool calls natively, while multi-turn and multi-step reasoning require further adaptation \cite{hf_functiongemma_modelcard}. Our work builds directly on this interface, extending it to Arabic and incorporating reasoning supervision to improve semantic disambiguation prior to tool invocation.

Recent research demonstrates that tool-calling performance does not transfer uniformly across languages. MASSIVE-Agents reformats a multilingual dataset into a BFCL-style function-calling benchmark spanning 52 languages and reports substantial cross-lingual disparities in correctness \cite{kulkarni2025massiveagents}. Even strong multilingual models exhibit significant degradation outside English under standardized evaluation. These findings underscore that structured execution requires language-specific training signals and schema adaptation.

Beyond individual models, production agent systems require orchestration frameworks capable of tool routing, state management, and multi-step execution. AutoGen provides a programmable multi-agent conversation framework where agents collaborate with tools and humans-in-the-loop \cite{wu2024autogen}. AgentBench evaluates LLMs as agents across interactive environments, emphasizing reasoning and decision-making under multi-turn conditions \cite{liu2023agentbench}. Frameworks such as LangChain/LangGraph and CrewAI operationalize stateful graphs and multi-agent workflows in open ecosystems, though governance and evaluation discipline are often left to implementers. The Model Context Protocol (MCP) proposes interoperability standards for connecting models to tools and data sources, including explicit safety considerations \cite{anthropic_mcp_2024,mcp_spec_2025}. These developments highlight the necessity of architectural abstractions that treat telemetry, policy enforcement, and evaluation as first-class concerns rather than post-hoc additions.

As agentic systems transition from research prototypes to real-world deployment, governance and risk management become central. The NIST AI Risk Management Framework (AI RMF 1.0) formalizes lifecycle risk governance for AI systems \cite{nist2023airmf}. ISO standards such as ISO/IEC 42001 and ISO/IEC 23894 define organizational controls for AI management and risk mitigation \cite{iso42001_2023,iso23894_2023}. Observability frameworks like OpenTelemetry provide standardized tracing mechanisms that can instrument agent execution flows in production environments \cite{opentelemetry_traces_2026}.

AISA (Agentic AI Systems Architecture) proposes a layered reference architecture that elevates evaluation, deployment, and governance as core system components rather than peripheral engineering tasks \cite{nacar2026aisa}. Our work operationalizes these architectural principles in the context of Arabic tool calling, demonstrating how localized datasets and models can be integrated within structured governance-aware pipelines.

\section{Methodology}

We describe the end-to-end pipeline used to construct, repair, and fine-tune an Arabic function-calling model based on \texttt{unsloth/functiongemma-270m-it}~\footnote{\url{https://huggingface.co/unsloth/functiongemma-270m-it}}. The methodology follows a data-centric and systems-aware approach: (i) structural auditing and schema repair of the dataset, (ii) prompt-length reduction via tool sampling, (iii) format-aligned chat construction compatible with FunctionGemma control tokens, and (iv) full-parameter supervised fine-tuning under completion-only masking.

\subsection{Model Foundation}

We initialize from \texttt{unsloth/functiongemma-270m-it}, a 270M-parameter variant of FunctionGemma optimized for structured function calling. Unlike parameter-efficient fine-tuning (e.g., LoRA), we adopt \textbf{full fine-tuning}, updating all parameters:

\begin{equation}
\theta^{*} = \arg\min_{\theta} \; \mathbb{E}_{(x,y)\sim \mathcal{D}} \left[ - \log P_{\theta}(y \mid x) \right],
\end{equation}

where $x$ is the formatted chat prompt (developer turn + tool declarations + user query), $y$ is the assistant’s structured function call, and $\mathcal{D}$ is the curated Arabic function-calling dataset. Completion-only masking ensures gradients are computed only over the assistant’s function-call tokens.

\subsection{Dataset Auditing and Structural Repair}

The \textbf{Arabic Function Calling} dataset\footnote{\url{https://huggingface.co/datasets/HeshamHaroon/Arabic_Function_Calling}} serves as the primary data source for this study. It comprises 50,810 samples spanning 36 tools, five major Arabic dialects (MSA, Egyptian, Gulf, Levantine, and Maghrebi), and eight real-world domains. While the dataset provides broad dialectal and functional coverage, preliminary fine-tuning experiments revealed several structural limitations that materially affected training stability and evaluation reliability.

\begin{figure}[t]
\centering
\includegraphics[width=\linewidth]{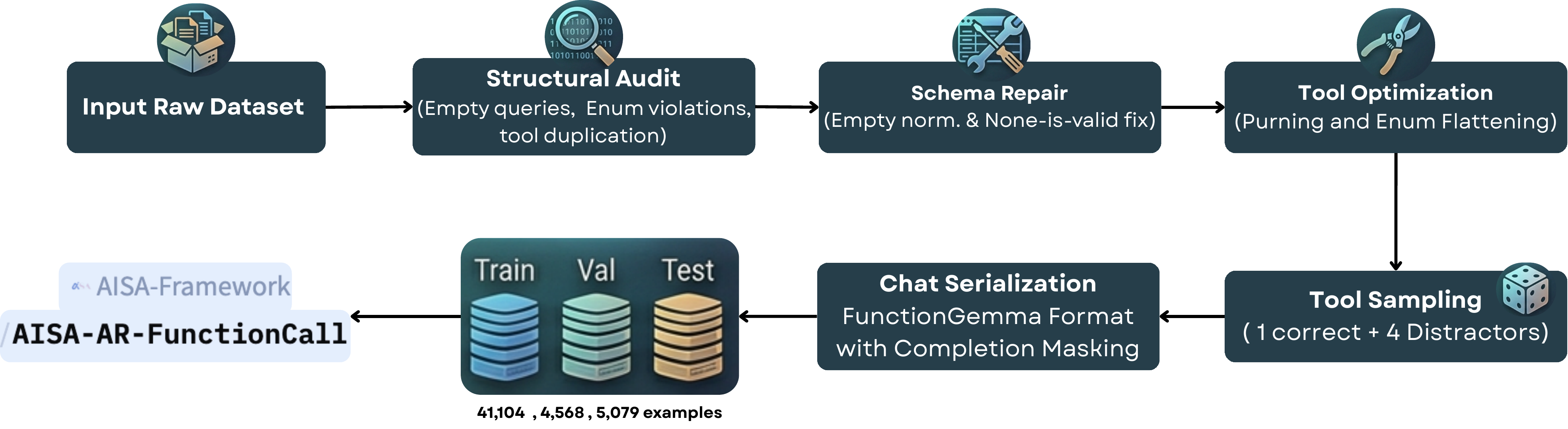}
\caption{
End-to-end transformation pipeline for AISA-AR-FunctionCall.
The process includes structural auditing, schema repair, tool optimization,
stochastic tool sampling, chat serialization using the FunctionGemma format,
and stratified train/validation/test splitting.
}
\label{fig:aisa_pipeline}
\end{figure}

Figure~\ref{fig:aisa_pipeline} presents the complete transformation pipeline adopted in this work. As illustrated, the process begins with a structural audit phase, where empty queries, enum violations, and duplicated tool definitions are identified. This is followed by schema repair, including normalization of argument values and correction of enum constraints (notably the \textit{None-is-valid} fix for optional parameters). Next, tool optimization is performed through pruning of unstable or redundant tools and flattening of enum constraints into descriptive fields to reduce schema rigidity. These steps collectively stabilize the supervision signal before downstream prompt construction and model training.

This staged repair process converts the original raw dataset into \textbf{AISA-AR-FunctionCall}, a schema-consistent and production-ready corpus specifically engineered for structured function-calling fine-tuning.

Empirical inspection revealed four dominant failure modes: (i) silent outputs for negative samples, (ii) enum constraint violations, (iii) duplicated or semantically overlapping tool definitions, and (iv) systematic prompt truncation caused by excessive tool declarations. These deficiencies directly impaired supervision quality and necessitated structured repair prior to model optimization.

\paragraph{Enum Compliance Correction.}
A critical source of data loss was the validation logic applied to enum-constrained parameters. The original filtering rule considered a sample valid only if the parameter value $v$ belonged to the predefined enum set:
\begin{equation}
\text{valid} = (v \in \text{Enum}),
\end{equation}
thereby incorrectly treating \texttt{None} values—used to indicate optional or unspecified arguments—as violations. This resulted in systematic exclusion of otherwise valid samples, particularly for tools with optional enum fields. The validation rule was corrected to explicitly allow null assignments:
\begin{equation}
\text{valid} = (v = \texttt{None}) \;\lor\; (v \in \text{Enum}),
\end{equation}
ensuring that absence of a value is interpreted as permissible when the parameter is not required. This modification restored thousands of previously discarded training instances and reactivated six tools that had effectively become “dead” due to complete sample exclusion.

\paragraph{Enum Normalization and Tool Pruning.}
In addition to correcting enum validation logic, we performed systematic normalization of enum values and structural consolidation of the tool inventory. Several enum-constrained parameters contained heterogeneous representations, including Arabic surface forms, variant English spellings, or free-text values that did not match the canonical schema. To enforce consistent supervision signals, these variants were mapped to standardized enum values defined in the tool schema, thereby reducing label fragmentation and improving alignment between arguments and schema definitions. 

Concurrently, an error-driven audit revealed that a subset of tools contributed disproportionate noise due to severe schema inconsistencies, duplicated functional intent, or unstable argument structures. For example, duplicated currency-conversion tools and overlapping time-retrieval functions fragmented the learning signal across semantically equivalent operations. After consolidation and removal of high-noise tools, the effective tool inventory was reduced from 36 to 27 tools. This pruning step decreased schema variability, reduced parameter-type violations, and stabilized supervision across domains, resulting in a more coherent and learnable action space for fine-tuning.

\subsection{Prompt Length Reduction via Tool Sampling}

A major bottleneck in early training experiments was prompt truncation. When all available tool declarations were included in every training instance (originally 36 tools, later 27 after pruning), the serialized chat prompt frequently exceeded 4,900 tokens. Given a maximum sequence length of 2048 tokens, this resulted in systematic truncation before the assistant’s function-call response, effectively preventing the model from observing the supervision signal. Consequently, gradient updates were dominated by prompt tokens rather than the structured action output.

To address this issue, we introduce a \textbf{stochastic tool sampling} strategy that constrains each training instance to a fixed-size subset of tools. Each example contains exactly five tool declarations. For positive samples (i.e., \texttt{requires\_function=True}), the sampled set consists of the ground-truth tool $t^{*}$ and four randomly selected distractor tools drawn without replacement from the remaining tool inventory. For negative samples, five tools are sampled uniformly at random, with no designated correct tool. The final subset is randomly permuted to eliminate positional bias.

Formally, for a positive training instance $i$ with correct tool $t^{*}$ and full tool inventory $\mathcal{T}$, the sampled tool set is:

\begin{equation}
\mathcal{T}_i = \pi \left( \{ t^{*} \} \cup \text{Sample}(\mathcal{T} \setminus \{t^{*}\}, 4) \right),
\end{equation}

where $\pi(\cdot)$ denotes a random permutation operator. For negative instances:

\begin{equation}
\mathcal{T}_i = \pi \left( \text{Sample}(\mathcal{T}, 5) \right).
\end{equation}

This mechanism reduces the median prompt length from approximately 4,900 tokens to approximately 793 tokens, ensuring all examples fit within the 2048-token context window. Beyond preventing truncation, stochastic sampling introduces implicit data augmentation: across epochs, each instance is paired with different distractor combinations, encouraging the model to discriminate the correct tool under varying contextual alternatives. Empirically, this substantially improves stability and convergence in lightweight (270M parameter) structured function-calling models. After tool subset construction, each instance is serialized into the FunctionGemma-compatible chat format described below.

\begin{algorithm}[t]
\caption{Stochastic Tool Sampling for Structured Function Calling}
\label{alg:tool_sampling}
\begin{algorithmic}[1]
\Require Full tool inventory $\mathcal{T}$, 
        ground-truth tool $t^{*}$ (optional), 
        flag $requires\_function$
\Ensure Sampled tool subset $\mathcal{S}$ of size 5

\If{$requires\_function = \textbf{True}$}
    \State $\mathcal{D} \gets \mathcal{T} \setminus \{t^{*}\}$
    \State $\mathcal{R} \gets \text{UniformSample}(\mathcal{D}, 4)$
    \State $\mathcal{S} \gets \{t^{*}\} \cup \mathcal{R}$
\Else
    \State $\mathcal{S} \gets \text{UniformSample}(\mathcal{T}, 5)$ where $\text{UniformSample}(\cdot, k)$ denotes sampling without replacement
\EndIf
\State $\mathcal{S} \gets \text{RandomShuffle}(\mathcal{S})$
\State \Return $\mathcal{S}$
\end{algorithmic}
\end{algorithm}
\subsection{Chat Template Construction}

Each training instance is serialized using the native FunctionGemma control-token format to preserve structural alignment between tool declarations and assistant outputs. Concretely, every example follows a four-part structure: (i) a developer turn containing the system instruction and a dynamically injected timestamp (to support temporal reasoning for expressions such as “tomorrow” or “next Monday”), (ii) a sampled set of tool declarations encoded with control tokens, (iii) the user query in Arabic, and (iv) the assistant’s structured function call. 

Completion-only masking is enabled by specifying \texttt{dataset\_text\_field="text"} during training, which automatically masks all prompt tokens (developer, tool declarations, and user turns) and computes loss exclusively over the assistant’s function-call output. This ensures that gradients optimize structured action generation rather than prompt reproduction.

\begin{figure}[t]
\centering
\includegraphics[width=\linewidth]{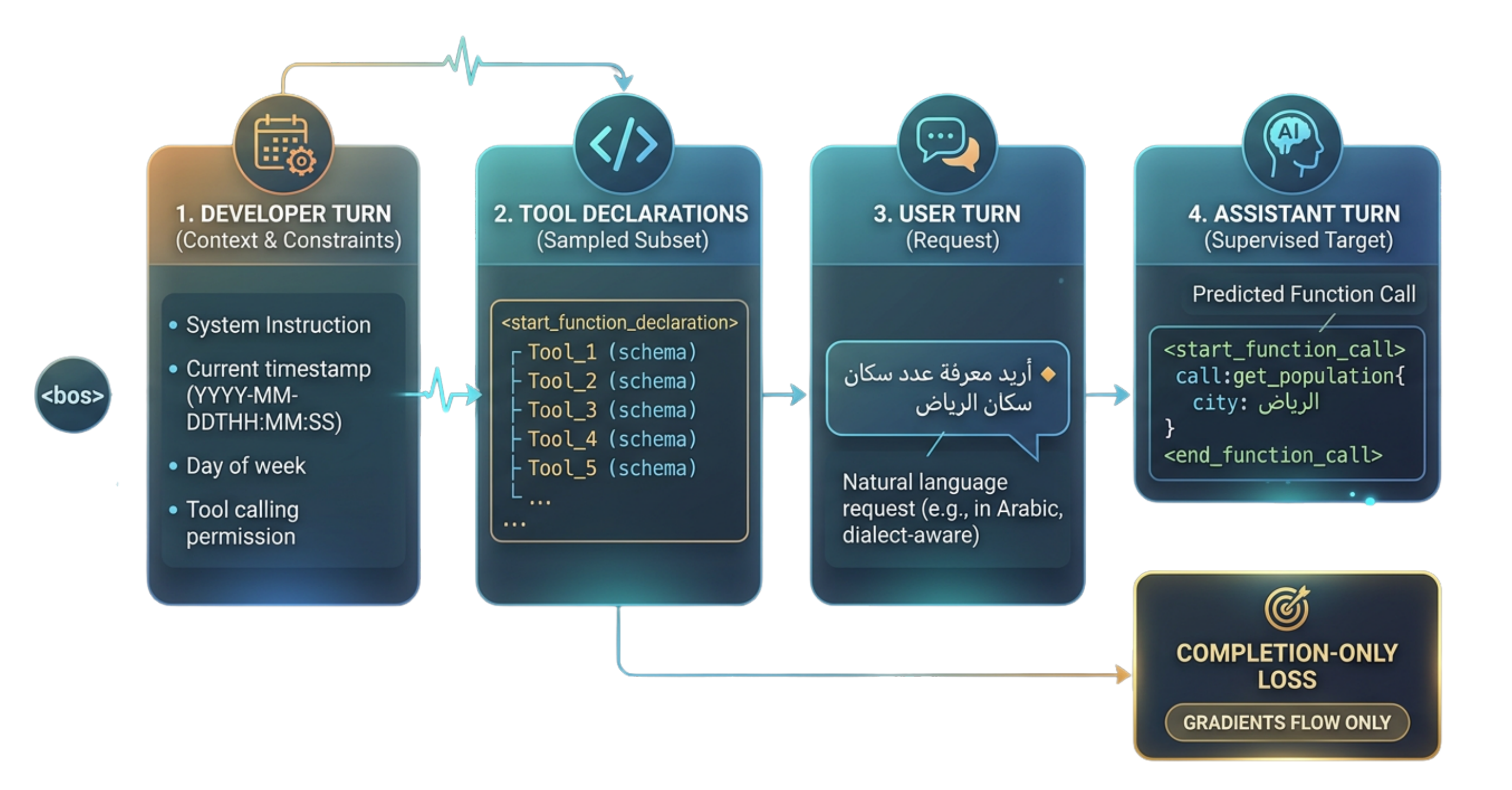}
\caption{Example of serialized training instance using the FunctionGemma control-token format.}
\label{fig:chat_serialization_pipeline}
\end{figure}

\subsection{Dataset Splitting and Training Configuration}

All experiments are conducted on \textbf{AISA-AR-FunctionCall}\footnote{\url{https://huggingface.co/datasets/AISA-Framework/AISA-AR-FunctionCall}}, a production-ready Arabic function-calling dataset released under the AISA framework. The dataset is fully schema-validated, tool-normalized, and formatted for direct fine-tuning of structured function-calling models such as FunctionGemma.

The data split follows the original metadata annotations to avoid distributional drift between training and evaluation. After formatting, tool sampling, and filtering, the corpus was partitioned into 41,104 training examples, 4,568 validation examples, and a held-out test set of 5,079 examples. The split preserves domain and dialect distributions across partitions, and stratification is applied to prevent tool or domain leakage. The test set remains strictly unseen during training and hyperparameter tuning.

Training is performed via full-parameter fine-tuning of all 268M model weights. We train for two epochs using a per-device batch size of 4 and gradient accumulation of 8, resulting in an effective batch size of 32. Optimization is carried out using 8-bit AdamW with a cosine learning rate scheduler and an initial learning rate of $2 \times 10^{-5}$. Gradient checkpointing is enabled to reduce memory overhead during backpropagation, enabling stable full-parameter training within hardware constraints. This configuration provides a balanced trade-off between convergence stability and computational efficiency for lightweight structured execution models.

\subsection{Full-Parameter Fine-Tuning Protocol}

We adopt full-parameter supervised fine-tuning of the 270M-parameter FunctionGemma model, updating all trainable weights rather than using parameter-efficient adaptation methods (e.g., LoRA). Let $\theta$ denote the full parameter set of the model. Given a serialized training instance $x$ and structured function-call target $y$, optimization follows the standard causal language modeling objective:

\begin{equation}
\mathcal{L}(\theta) = - \sum_{t \in \mathcal{Y}} \log P_{\theta}(y_t \mid x, y_{<t}),
\end{equation}

where $\mathcal{Y}$ indexes only the assistant tokens corresponding to the function-call output. 

To prevent prompt-token dominance during training, we employ completion-only masking. Let the serialized sequence be decomposed as:

\begin{equation}
x = [x_{\text{dev}}, x_{\text{tools}}, x_{\text{user}}, y_{\text{assistant}}],
\end{equation}

where the first three segments constitute the prompt and $y_{\text{assistant}}$ represents the supervised function call. A binary mask $m_t$ is applied such that:

\begin{equation}
m_t =
\begin{cases}
0 & \text{if } t \in \{x_{\text{dev}}, x_{\text{tools}}, x_{\text{user}}\}, \\
1 & \text{if } t \in y_{\text{assistant}}.
\end{cases}
\end{equation}

The loss is then computed as:

\begin{equation}
\mathcal{L}(\theta) = - \sum_{t} m_t \log P_{\theta}(x_t \mid x_{<t}).
\end{equation}

This ensures gradients are propagated exclusively through structured action tokens rather than through prompt content, thereby concentrating optimization on executable function generation. All 268M parameters are updated during training:

\begin{equation}
\theta \leftarrow \theta - \eta \nabla_{\theta} \mathcal{L}(\theta),
\end{equation}

where $\eta$ denotes the learning rate. Unlike adapter-based approaches, this strategy allows the model to fully realign internal representations toward Arabic structured execution rather than merely learning lightweight projection layers.

Training is conducted in BF16 precision with gradient checkpointing enabled to reduce memory footprint. Gradient accumulation is used to simulate a larger effective batch size without exceeding hardware constraints. Additionally, 8-bit AdamW optimization is employed to reduce optimizer-state memory overhead while preserving convergence stability. Preliminary experiments with partial adaptation indicated instability in function-name prediction and argument alignment, particularly under dialectal variation. Full-parameter fine-tuning yielded significantly improved convergence behavior and more stable structured outputs, suggesting that deeper representational adaptation is required for reliable Arabic function calling in lightweight models.

\subsection{Reasoning-Augmented Fine-Tuning (Exploratory Variant)}

In addition to the primary full fine-tuning model described above, we conduct an exploratory reasoning-augmented fine-tuning experiment to evaluate whether explicit reasoning supervision improves structured tool invocation behavior. This variant builds upon the fully fine-tuned AISA-AR-FunctionCall model and introduces intermediate reasoning tokens prior to tool execution. For a subset of the dataset (12k samples), we augment assistant outputs with explicit reasoning segments enclosed within \texttt{<think>} and \texttt{</think>} tags, followed by the structured tool call. The modified generation pattern becomes:

\begin{equation}
x = [x_{\text{prompt}}, \texttt{<think>} \; r \; \texttt{</think>}, \texttt{<start\_function\_call>} \dots ]
\end{equation}

where $r$ denotes a short reasoning trace explaining tool selection and argument extraction. During inference, the model is primed to begin its turn with \texttt{<think>\textbackslash n}, ensuring that reasoning cannot be skipped.

Unlike the primary production model, which employs full-parameter fine-tuning, the reasoning-augmented variant is trained using LoRA adaptation with increased capacity. Specifically, we use a LoRA rank of $r = 64$ (increased from 16), $\alpha = 64$, and dropout of 0.05, resulting in approximately 5.36\% trainable parameters. This configuration enables targeted behavioral adaptation while preserving the underlying base model weights. Completion-only masking is retained, such that gradients propagate exclusively through the reasoning (\texttt{<think>}) segment and the subsequent structured function-call tokens:
\begin{equation}
\mathcal{L}(\theta) =
- \sum_{t \in \{\text{think}, \text{tool\_call}\}}
\log P_{\theta}(x_t \mid x_{<t}).
\end{equation}

To mitigate hallucinated tool invocation, additional no-tool examples were incorporated during reasoning fine-tuning, explicitly supervising abstention when \texttt{requires\_function=False}. The reasoning model is trained for three epochs using a learning rate of $3 \times 10^{-6}$, an effective batch size of 32, cosine scheduling, and 8-bit AdamW optimization. The resulting model, \textbf{AISA-AR-FunctionCall-Think}, is released publicly\footnote{\url{https://huggingface.co/AISA-Framework/AISA-AR-FunctionCall-Think}} and is presented as an exploratory extension for structured reasoning analysis. The fully fine-tuned \textbf{AISA-AR-FunctionCall-FT} model remains the primary production-ready system.
\section{Experiments and Results}

We evaluate three model variants: (i) the \textbf{Baseline}, a pre-finetuned FunctionGemma model without Arabic adaptation; (ii) \textbf{AISA-AR-FunctionCall-FT}, the fully fine-tuned production model; and (iii) \textbf{AISA-AR-FunctionCall-Think}, a reasoning-augmented LoRA variant. All evaluations are conducted on the held-out test set ($n=5079$).

\begin{figure}[!t]
\centering
\includegraphics[width=0.9\linewidth]{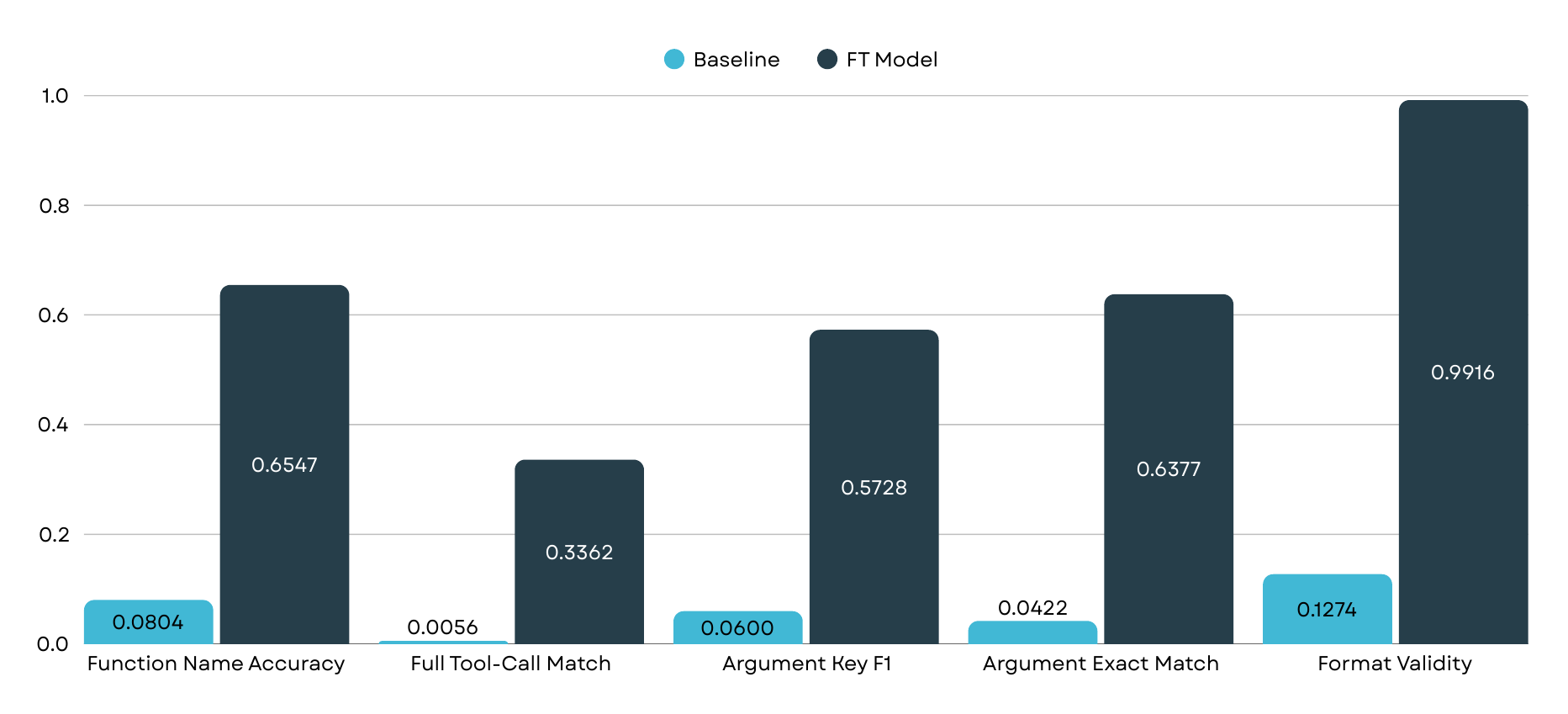}
\caption{
Core structured performance comparison between the baseline and the fully fine-tuned model.
Metrics include function selection accuracy, full tool-call match, argument alignment (F1 and exact match), and format validity.
}
\label{fig:core_metrics}
\end{figure}

\begin{figure}[!t]
\centering
\includegraphics[width=0.75\linewidth]{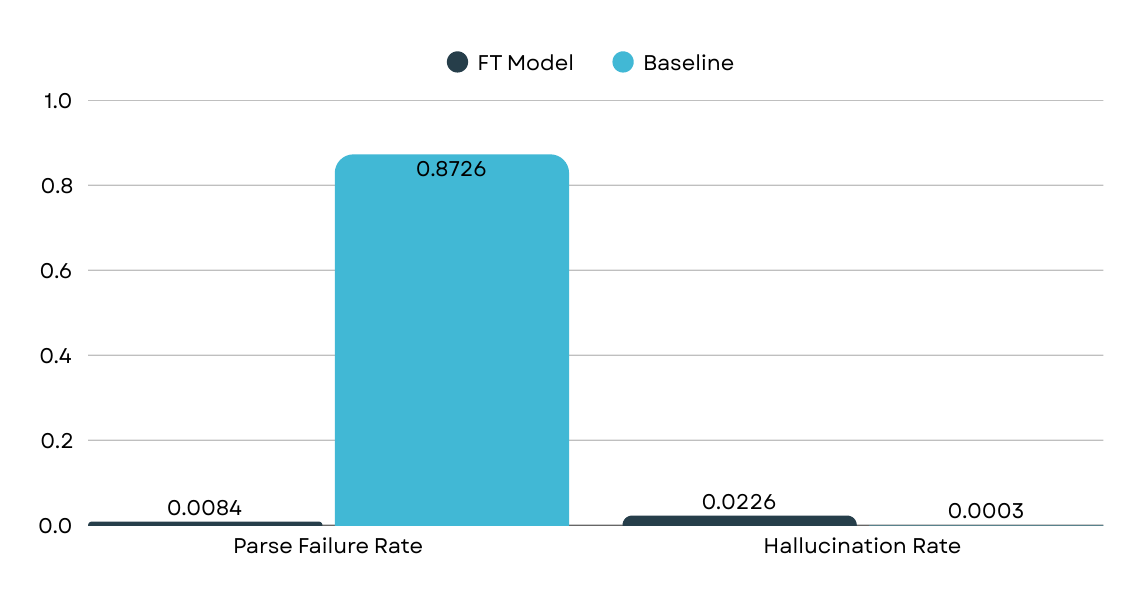}
\caption{
Structural stability comparison between the baseline and the fully fine-tuned model.
Metrics include parse failure rate and hallucination rate.
}
\label{fig:stability_metrics}
\end{figure}

Figures~\ref{fig:core_metrics} and~\ref{fig:stability_metrics} present a direct comparison between the baseline model and the fully fine-tuned AISA-AR-FunctionCall-FT model on clean positive samples ($n=2873$). This evaluation isolates cases where a function call is required and no enum violations are present, allowing structured execution reliability to be assessed under controlled conditions.

The baseline model exhibits systemic structural failure. More than 87\% of outputs fail to produce a valid structured function call, and full tool-call matches are effectively negligible. Tool selection accuracy remains below 8\%, confirming that multilingual pretraining alone is insufficient for reliable Arabic structured execution.

In contrast, full fine-tuning produces a decisive structural recovery. Parse failures drop from 87\% to below 1\%, and format validity approaches 100\%. Function name accuracy improves by more than eightfold, while argument alignment metrics show substantial gains across both key-level and exact-value evaluations. These results indicate that structured dataset repair, schema normalization, tool-aware sampling, and completion-only masking collectively restore stable and reliable Arabic function-calling behavior in a lightweight 270M-parameter model.

Importantly, the fine-tuned model preserves perfect abstention behavior on negative samples, maintaining 100\% accuracy when \texttt{requires\_function=False}, indicating that structural recovery does not come at the cost of increased over-calling.

To examine robustness across linguistic variation, Table~\ref{tab:dialect_performance} reports function name accuracy across five Arabic dialect groups. This comparison evaluates whether structured fine-tuning reduces dialectal performance gaps observed in the baseline model.

\begin{table}[!ht]
\centering
\caption{Function Name Accuracy by Dialect}
\label{tab:dialect_performance}
\begin{tabular}{lcc}
\toprule
Dialect & Baseline & FT Model \\
\midrule
MSA        & 0.0862 & \textbf{0.7613} \\
Gulf       & 0.0526 & \textbf{0.6972} \\
Egyptian   & 0.0493 & \textbf{0.6834} \\
Levantine  & 0.0633 & \textbf{0.6948} \\
Maghrebi   & 0.0452 & \textbf{0.6158} \\
\bottomrule
\end{tabular}
\end{table}

The baseline model exhibits consistently weak performance across all dialects, with accuracy remaining below 9\% even in MSA. This indicates that multilingual pretraining alone does not provide reliable structured execution in Arabic. Following full fine-tuning, performance improves dramatically across every dialect. Accuracy exceeds 68\% for all major dialect groups and reaches over 76\% in MSA. Importantly, the disparity between dialects narrows substantially compared to the baseline. This suggests that structured supervision and schema-aligned training reduce dialectal execution bias rather than amplifying it, resulting in more linguistically robust tool invocation behavior.

\begin{figure}[!ht]
\centering
\includegraphics[width=0.7\linewidth]{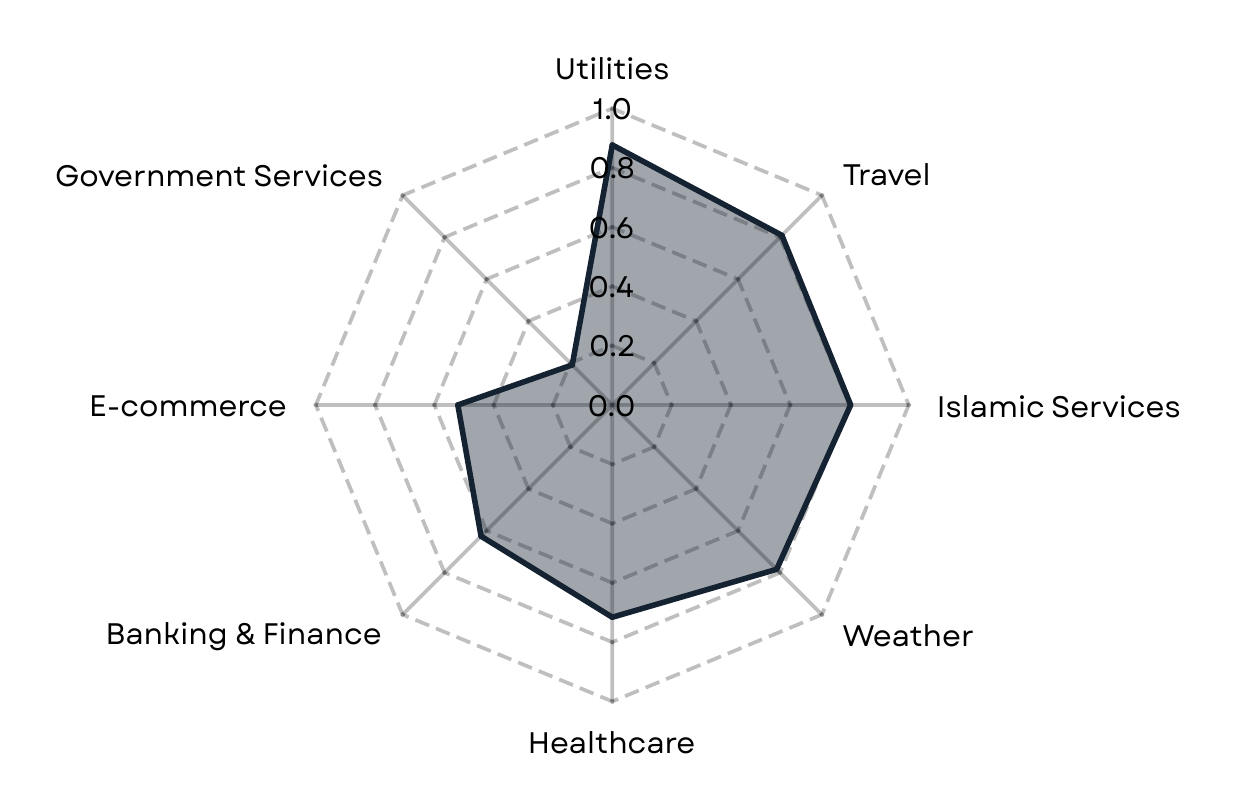}
\caption{
Function Name Accuracy by Domain for the fully fine-tuned 
\textbf{AISA-AR-FunctionCall-FT} model. 
Highly structured transactional domains (Utilities, Travel, Islamic Services, Weather) 
achieve strong performance, while procedurally complex domains such as 
Government Services remain more challenging.
}
\label{fig:domain_performance}
\end{figure}

Moreover, to assess task-specific robustness, Figure~\ref{fig:domain_performance} reports function name accuracy across the eight primary domains in the AISA-AR-FunctionCall dataset. This analysis focuses on the fine-tuned production model, as baseline performance is uniformly low across domains and primarily constrained by structural failure.

As shown in Figure~\ref{fig:domain_performance}, performance is strongest in highly structured transactional domains such as utilities, travel, weather, and Islamic services, where tool invocation patterns are relatively deterministic and argument schemas are well-defined. In contrast, domains involving regulatory or procedural complexity—most notably government services—exhibit substantially lower accuracy. Importantly, parse failure remains negligible across all domains, indicating that errors are primarily semantic (tool selection ambiguity or argument misalignment) rather than structural. These results suggest that remaining limitations arise from decision-level reasoning challenges rather than serialization instability.

\subsection{Failure Mode Analysis}

To better understand how model behavior evolves after fine-tuning, Table~\ref{tab:failure_shift} compares the distribution of error types between the baseline and the fully fine-tuned model.

\begin{table}[!ht]
\centering
\caption{Error Distribution Shift}
\label{tab:failure_shift}
\begin{tabular}{lcc}
\toprule
Error Type & Baseline (\%) & FT Model (\%) \\
\midrule
Parse Failure      & 80.7 & \textbf{0.76} \\
Tool Hallucination & \textbf{0.04} & 24.7 \\
Wrong Function     & \textbf{3.3}  & 23.6 \\
Argument Mismatch  & \textbf{5.4}  & 20.2 \\
Correct            & 0.3  & \textbf{20.3} \\
\bottomrule
\end{tabular}
\end{table}

The baseline model is overwhelmingly dominated by structural collapse, with more than 80\% of errors arising from parse failures. In this regime, the model frequently fails to emit a valid function-call structure, rendering semantic evaluation largely irrelevant.

After fine-tuning, parse failures are nearly eliminated. The error distribution shifts from structural failure to semantic misalignment. The majority of remaining errors involve incorrect tool selection, hallucinated tool invocation, or argument mismatches. This shift is significant: it indicates that structured serialization and schema alignment have been successfully learned, and that remaining challenges primarily concern decision-level reasoning and tool disambiguation rather than formatting instability.

\subsection{Qualitative Error Analysis}

To complement the quantitative evaluation, we manually inspect representative prediction errors from the fine-tuned model. While structural failures are largely eliminated, several recurring semantic patterns emerge:

\begin{itemize}
\item \textbf{Weather vs Air Quality Confusion:} Queries requesting weather forecasts are occasionally mapped to \texttt{get\_air\_quality}, indicating semantic overlap in environmental terminology.

\item \textbf{Banking vs Currency Conversion Misrouting:} Financial transfer requests are sometimes predicted as \texttt{convert\_currency}, suggesting partial lexical matching on monetary expressions without full intent disambiguation.

\item \textbf{Government vs Healthcare Cross-Domain Confusion:} Certain government service queries (e.g., visa or residency checks) are misrouted to healthcare-related tools, reflecting cross-domain interference under ambiguous procedural language.

\item \textbf{Ambiguous Natural Language Queries:} Underspecified user inputs occasionally trigger tool hallucination or incomplete argument extraction, particularly when essential parameters are implied rather than explicitly stated.

\item \textbf{Argument-Level Drift:} In some cases, the correct tool is selected but argument values exhibit semantic drift (e.g., normalized date formats, lexical variations).
\end{itemize}

These examples reinforce the earlier failure-mode shift analysis: remaining errors are primarily semantic rather than structural. Tool disambiguation under lexical overlap and implicit intent remains the principal challenge for further improvement.

\subsection{Reasoning-Augmented Variant}

To investigate whether explicit reasoning supervision improves structured tool invocation, we train a LoRA-based variant that generates an intermediate \texttt{<think>} segment prior to emitting the function call. This reasoning block is supervised during training and enforced during inference, ensuring that tool selection is preceded by an explicit decision trace.

\begin{table}[!ht]
\centering
\caption{Reasoning Model Results (Strict Evaluation, $n=240$)}
\label{tab:reasoning_strict}
\begin{tabular}{lc}
\toprule
Metric & Score \\
\midrule
Tool Call Rate         & 0.992 \\
Think-Before-Call Rate & 1.000 \\
Function Name Accuracy & 0.992 \\
Argument F1            & 1.000 \\
Decision Accuracy      & 0.992 \\
Hallucination Rate     & 0.000 \\
\bottomrule
\end{tabular}
\end{table}

Under strict evaluation—where both reasoning presence and correct tool invocation are required—the reasoning model demonstrates near-perfect alignment. The model consistently emits a structured reasoning block prior to the tool call and achieves flawless argument extraction on a stratified evaluation subset of 240 samples.

It is important to note that strict formatting validators classify many reasoning outputs as parse failures because the serialized output now includes \texttt{<think>} tokens before the function-call marker. This does not reflect structural instability, but rather a difference in output serialization. Under deployment-aware evaluation—where reasoning segments are permitted—the model maintains near-perfect tool invocation correctness.

This reasoning-augmented model is presented as an exploratory extension to analyze structured reasoning behavior. The primary production-ready system remains the fully fine-tuned AISA-AR-FunctionCall-FT model.

\section{Discussion}

This work demonstrates that reliable Arabic function calling is not primarily a model-size limitation, but a data and supervision alignment problem. The baseline results reveal systemic structural collapse, with the majority of outputs failing to produce valid function-call formats. This confirms that multilingual pretraining alone does not guarantee executable structured behavior in morphologically rich and dialectally diverse languages such as Arabic.

The full fine-tuned AISA-AR-FunctionCall-FT model shows that structured dataset repair, schema normalization, and tool-aware sampling are sufficient to restore stable function-calling behavior within a lightweight 270M-parameter model. Parse failures are nearly eliminated, format validity approaches 100\%, and function name accuracy increases by more than eightfold. These improvements indicate that structural serialization learning can be reliably achieved when prompt construction and supervision are carefully engineered.

However, the failure mode analysis highlights a second-stage challenge: once structural collapse is resolved, remaining errors shift toward semantic misalignment. Tool hallucination, incorrect function selection, and argument mismatches become the dominant error types. This suggests that structured learning and decision-level reasoning are separable phenomena. While serialization stability can be enforced through format-aware training, accurate tool selection requires deeper semantic grounding and possibly contrastive or ranking-based supervision.

Dialect-level results further indicate that structured supervision reduces multilingual execution bias. After fine-tuning, performance disparities between dialects narrow substantially, suggesting that schema-aligned training promotes robustness across linguistic variation rather than amplifying language imbalance.

The reasoning-augmented variant provides additional insight. When explicit reasoning traces are supervised, the model achieves near-perfect structured alignment within the evaluated subset. This suggests that intermediate reasoning can improve tool selection consistency and argument extraction fidelity. Nevertheless, reasoning supervision alters output serialization and introduces deployment considerations regarding formatting validation. Consequently, while reasoning improves decision alignment, it must be carefully integrated into production pipelines.

Overall, the findings emphasize that production-grade multilingual tool calling requires a layered approach: structural reliability first, followed by semantic calibration and decision refinement. The AISA-AR-FunctionCall framework provides an empirical demonstration of this progression.

\section{Conclusion}

This paper presents AISA-AR-FunctionCall, a production-oriented Arabic function-calling framework built through systematic dataset auditing, schema repair, tool-aware prompt restructuring, and full-parameter fine-tuning. We demonstrate that reliable Arabic tool invocation can be achieved within a lightweight 270M-parameter model when structural supervision is carefully engineered. Baseline results reveal severe structural collapse under multilingual pretraining alone, while the fine-tuned model nearly eliminates parse failures and substantially improves function selection and argument alignment across dialects and domains. Our analysis further shows that once structural reliability is restored, remaining limitations shift toward semantic decision-level errors, such as incorrect tool disambiguation and argument mismatch. This suggests a two-stage progression for multilingual agentic systems: first ensuring format and schema stability, then improving semantic calibration. The reasoning-augmented variant provides additional evidence that explicit intermediate reasoning can enhance tool-selection alignment, although integration into production pipelines requires careful serialization management. Overall, the results highlight that multilingual structured execution is primarily a data and supervision alignment challenge rather than a model-scale limitation. The AISA-AR-FunctionCall framework offers both a production-ready training corpus and a research testbed for advancing structured tool use in Arabic and other morphologically rich languages. Future work will explore contrastive supervision, tool-ranking refinement, and confidence-based calibration to further close the semantic gap toward deployment-grade reliability.

\bibliographystyle{unsrt}
\bibliography{references}

\end{document}